\documentclass[runningheads]{llncs}
\usepackage[T1]{fontenc}
\usepackage{graphicx,verbatim}
\usepackage{booktabs, multicol, multirow}
\usepackage[table]{xcolor}
\usepackage{amsmath}
\usepackage{booktabs}
\usepackage{multirow}
\usepackage{pifont}
\usepackage{xurl}
\usepackage{bbding}
\usepackage{fontawesome5}
\usepackage[hidelinks]{hyperref}

\begin{document}

\title{FunPiQ: A New Benchmark for Pixel-Level Quality Assessment in Fundus Images}
\titlerunning{FunPiQ}

\author{
    Pengwei Wang        \inst{1}\textsuperscript{(\faEnvelope[regular])}      \and 
    Jos\'e Morano       \inst{1,2}    \and 
    Virginia Mares      \inst{1}      \and 
    Hrvoje Bogunovi\'c  \inst{1,2} 
    }
\authorrunning{P. Wang et al.}
\institute{
    Institute of Artificial Intelligence, Center for Medical Data Science, Medical University of Vienna, Vienna, Austria \and
    Christian Doppler Lab for Artificial Intelligence in Retina, Center for Medical Data Science, Medical University of Vienna, Vienna, Austria
    \email{\{pengwei.wang,jose.moranosanchez,hrvoje.bogunovic\}@meduniwien.ac.at}
    }
  
\maketitle

\begin{abstract}
Color fundus photography (CFP) is the most common ophthalmic imaging modality for large-scale screening. However, it is highly susceptible to degradations, making robust fundus image quality assessment (FIQA) crucial. The criteria for what constitutes high-quality at the image level vary across clinical tasks, making FIQA dependent on expert knowledge. This motivated the development of automated methods and datasets. While existing datasets aim to standardize image-level quality, their criteria often differ. Furthermore, image-level labels preclude the quantitative evaluation of localized degradations, which is essential for trustworthy FIQA. We argue that pixel-level FIQA based on anatomical visibility represents a more task-agnostic, explainable approach. In this work, we introduce FunPiQ, the first FIQA benchmark to provide pixel-level quality annotations. In addition, we propose EFIQA-CP, an explainable-by-design (EBD) method that uses quality pseudo-labels based on anatomical visibility to train a CNN via Non-Negative Positive-Unlabeled learning. Extensive evaluations of classification methods with post-hoc explanations, anomaly detection methods, and EBD methods demonstrate the superior performance of the last and, particularly, of EFIQA-CP. Our code, weights, and dataset are publicly available at \url{https://github.com/penway/FunPiQ}

\end{abstract}

\section{Introduction}

\begin{figure}[t]
    \centering
    \includegraphics[width=1\linewidth]{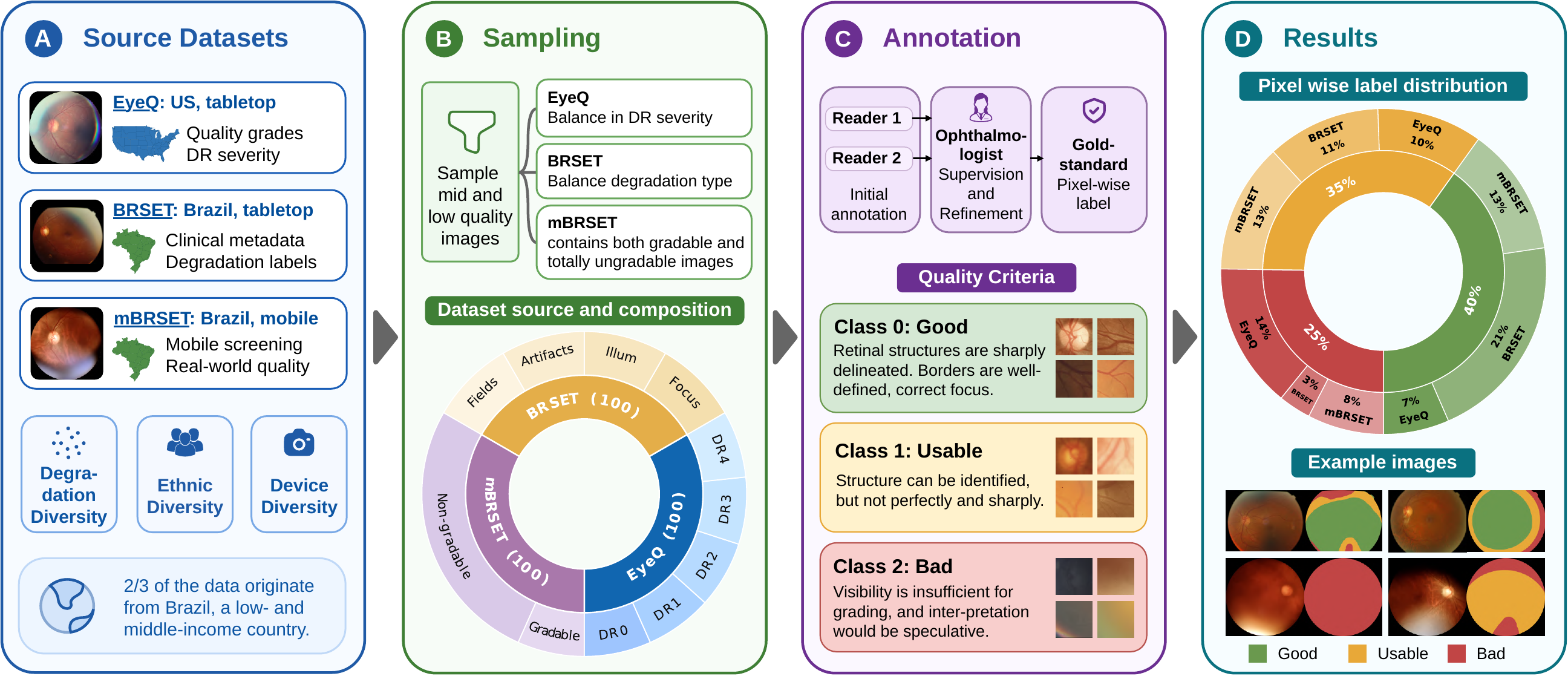}
    \caption{Overview of our FunPiQ dataset.}
    \label{fig:figabs}
\end{figure}

Color fundus photography (CFP) is one of the most widely used imaging modalities in ophthalmology~\cite{fogellevin2022advanced}, especially for screening. CFP enables the discovery not only of retinal diseases but also of systemic health conditions~\cite{zhu2025oculomics,grzybowski2024retina}. In addition, it is arguably the most accessible ophthalmic modality, with devices ranging from professional desktop systems to portable, mobile-phone-based cameras, making large-scale screening feasible in remote areas~\cite{panwar2016fundus}.

However, CFP is highly susceptible to degradations, such as inconsistent illumination and poor contrast~\cite{fu2019eyeq}. These issues often affect \textit{local} regions rather than the entire image, meaning certain areas remain of high quality while others are degraded (see Fig.~\ref{fig:figabs}). Such degradations are frequent even in professional settings. For example, in the EyePACS benchmark~\cite{KaggleDR}, 42\% of images had quality issues, with 19\% deemed unsuitable for diagnosis~\cite{fu2019eyeq}. These challenges are further exacerbated when using portable fundus cameras in uncontrolled environments. Thus, Fundus Image Quality Assessment (FIQA) represents a crucial step.

Nonetheless, FIQA faces important challenges, primarily due to the lack of standardized criteria. The definition of \emph{image-level} quality is inherently tied to specific downstream tasks, as the visibility of certain anatomical structures may be critical for diagnosing some diseases while remaining less relevant, or irrelevant, for others. This task-dependency makes manual, image-level FIQA partly subjective, time-consuming, and dependent on expert knowledge, rendering it unfeasible for large-scale screening. Consequently, there is a growing need for automated methods and datasets.

Table~\ref{tab:summary} summarizes the most relevant datasets for FIQA. While these benchmarks provide various quality-related labels, ranging from global classes to specific degradation types, they implement distinct criteria for what constitutes ``quality''. As previously discussed, a universal, image-level standard remains elusive because diagnostic requirements vary by disease; consequently, some images labeled as \textit{bad} quality in these datasets are actually usable for specific clinical tasks. For example, in MSHF~\cite{jin2023mshf}, images flagged as \textit{bad} quality were still deemed suitable for glaucoma screening. Furthermore, the separate standardization efforts have led to fairly dissimilar classifications across benchmarks, evidenced by the suboptimal generalization of models on external tests~\cite{wang2026efiqa},
and by differences in quality labels assigned to the same images by independent teams---for instance, an 11\% difference in a subset of EyePACS~\cite{fu2019eyeq,zhou2018eyepacsq}.
Given the inherently local nature of degradations, we argue that defining quality locally, through the visibility of underlying anatomy, offers a more task-agnostic, explainable alternative. This approach can be coupled with specific anatomical requirements (e.g., optic disc visibility) to implement dynamic image-level quality criteria.

\begin{table}[tb]
    \centering
    \caption{Summary of current CFP datasets with quality labels, including classes (C) and degradation types (DT). 2C includes \textit{good} and \textit{bad}, while 3C adds \textit{usable}.
    FunPiQ is the first evaluation dataset providing pixel-level quality annotations.
    }
    \label{tab:summary}
    \setlength{\tabcolsep}{8pt} 
    \renewcommand{\arraystretch}{1.2}
    \resizebox{\textwidth}{!}{%
        \begin{tabular}{clllll}
            \toprule
            \textbf{Granularity} & \textbf{Dataset} & \textbf{Year} & \textbf{Size} & \textbf{Country} & \textbf{Quality Labels} \\ \midrule
            \multirow{8}{*}{\textbf{Image-level}}
            & DRIMDB \cite{csevik2014drimdb}    & 2014 & 216    & Turkey  & 2C and \textit{outlier} class \\
            & EyePACS-Q \cite{zhou2018eyepacsq} & 2018 & 88,702 & US      & 2C \\
            & EyeQ \cite{fu2019eyeq}            & 2019 & 30,000 & US      & 3C \\
            & DeepDRiD \cite{liu2022deepdrid}   & 2022 & 2,256  & China   & 2C and 0--10 score with 3DT \\
            & MSHF \cite{jin2023mshf}           & 2023 & 1,302  & China   & 2C and 3DT \\
            & BRSET \cite{nakayama2024brset}    & 2024 & 16,266 & Brazil  & 2C and 4DT \\
            & mBRSET \cite{wu2025mbrset}        & 2025 & 5,164  & Brazil  & 2C \\
            & FQS \cite{gong2025fqs}            & 2025 & 2,246  & China   & 3C and mean opinion score \\ \midrule
            \rowcolor{gray!10}
            \textbf{Pixel-level}
            & \textit{\textbf{FunPiQ}}                   & 2026 & 300    & US, Brazil & 3C \\
            \bottomrule
        \end{tabular}%
    }
\end{table}

Most FIQA methods rely on supervised deep learning on the aforementioned datasets, incorporating various architectural innovations. For instance, MCF-Net utilizes multi-color space fusion~\cite{fu2019eyeq}; FGR-Net, reconstruction-based representations~\cite{khalid2024fgr}, and Shen et al. use semi-tied adversarial domain adaptation for domain-invariant features~\cite{shen2020domain}.
However, these approaches depend on image-level labels, inheriting the biases of their training datasets. Furthermore, while post-hoc visualization techniques like Grad-CAM~\cite{selvaraju2017grad} are frequently used~\cite{shen2020domain,khalid2024fgr}, the explainability provided by these tools is limited, and the localization of degradations is generally imprecise. To address the latter, EyeQual~\cite{costa2017eyequal} proposes an explainable-by-design architecture that outputs local-level quality maps via patch-level scoring. Aiming to address both issues, EFIQA~\cite{wang2026efiqa} proposed using vessel unsupervised anomaly detection (VUAD) as a prior for producing local-level quality maps without the need for quality-related labels. While it showed superior performance on external tests, EFIQA still presents some limitations.
First, it relies on a single linear adapter for patch-level quality scores from foundation model features, with no global context. This leads to incorrect predictions in areas with naturally low vessel density, such as the fovea.
Second, it is trained on the noisy pseudo-labels from the VUAD model via a simple Binary Cross-Entropy (BCE) loss, leading to pseudo-label overfitting.
Finally, despite the recognized importance of explainability and localization~\cite{khalid2024fgr,shen2020domain}, no benchmark currently exists to quantitatively evaluate the precision of these outputs.

In this context, the contributions of this work are threefold:
\begin{enumerate}
    \item We introduce the Fundus Pixel Quality (FunPiQ) dataset, the first public benchmark dedicated to pixel-level quality assessment for CFP. The benchmark comprises 300 images sourced from three public datasets and annotated by expert readers supervised by a board-certified ophthalmologist.
    \item We propose EFIQA-CP, a method that mitigates EFIQA's limitations by using a shallow CNN with a large receptive field as feature adapter and Non-Negative Positive-Unlabeled (nnPU) learning~\cite{kiryo2017nnpu} to handle pseudo-labels.
    \item We conduct a thorough evaluation of existing and proposed methods on our benchmark, demonstrating the superior performance of EFIQA-CP and highlighting the potential of explainable-by-design approaches for implementing robust, local-level quality assessment systems.
\end{enumerate}

\section{Materials and Methods}

\subsection{Dataset}

\paragraph{Data Selection.}
Fig.~\ref{fig:figabs} shows an overview of the dataset.
To maximize data diversity in terms of degradation types, ethnicity, and acquisition devices, we sampled from three existing benchmarks from two different countries: EyeQ~\cite{fu2019eyeq} (US), BRSET~\cite{nakayama2024brset,BRSET-physio,Goldberger2000PhysioNet}, and mBRSET~\cite{wu2025mbrset,mBRSET-physio,Goldberger2000PhysioNet} (both Brazil). Thus, 2/3 of the data come from a low- and middle-income country (LMIC), typically underrepresented in benchmarks. Moreover, these datasets include images captured using a wide range of hardware, from high-end tabletop fundus cameras to mobile phone-based portable devices.
Since FunPiQ is designed to evaluate the localization of degradations by IQA methods (i.e. their explainability) task-agnostically (not mirroring specific prevalences), for all three datasets, we selected the images labeled as \textit{bad}. This produces a mix of good, usable, and bad regions within each image (Fig.~\ref{fig:figabs} D), so that every sample tests a model's ability to differentiate quality locally.
For EyeQ, we balanced the dataset by selecting all types of DR level. For BRSET, we balanced among degradation types (illumination, focus, field, and artifact). For mBRSET, some images have diagnosis, while others are not gradable. We sampled all the gradable ones, and randomly sampled the rest.

\paragraph{Labeling.}
Images were annotated by expert readers and supervised by a board-certified ophthalmologist. Quality was assessed at the pixel level across the entire fundus image, based on the visibility and interpretability of anatomical or pathological features. Three quality categories were defined: \textit{good} (regions of optimal visibility with sharply delineated retinal features and no relevant artifacts), \textit{usable} (regions with mild reduction in sharpness or contrast without critical impact on structural identification), and \textit{bad} (non-gradable regions due to severe artifacts, signal dropout, shadowing, media opacity, or marked defocus).

\subsection{Method}
Our method follows a similar overall pipeline to EFIQA~\cite{wang2026efiqa}, and utilizes its VUAD model to generate local quality pseudo-labels. These labels are then used to train an adapter that maps features from a pretrained foundation model to final quality maps. However, these pseudo-labels are inherently noisy, as they rely solely on the absence of visible vessels. In EFIQA, the adapter is a linear layer that takes the features of a single patch, with no global context, and outputs its quality score. This setup, together with the noisy pseudo-labels, causes the network to flag as bad quality regions with naturally low vessel density (e.g., the macula). To mitigate these issues, we propose a wide-yet-shallow network that integrates spatial information through a broader receptive field. Furthermore, we employ nnPU learning to address the label distribution problem and improve robustness against the noise of the pseudo-labels.

\paragraph{Adapter Architecture.}
We extract features from a frozen DINOv3 backbone \cite{dinov3}. The network processes these features through five ConvNeXt-style blocks utilizing $7\times7$ depthwise convolutions, GELU activations \cite{hendrycks2016gelu}, and channel-wise Layer Normalization \cite{ba2016layernorm}. Different from standard setting, we use the MLP ratio as 1 instead of 4 to limit the capacity of the model and avoid overfitting.

\paragraph{nnPU.}
To handle noisy pseudo-labels, we apply a strict threshold to isolate reliable low-quality pixels, treating the rest as unlabeled. This motivates a Positive-Unlabeled (PU) learning approach, which effectively learns an unbiased classifier without requiring guaranteed high-quality (negative) labels. Furthermore, because highly flexible neural networks can overfit in this setup and push the estimated risk of the hidden negatives below zero, we utilize the non-negative PU (nnPU) framework~\cite{kiryo2017nnpu} to clamp this risk at zero, stabilizing training. Using a softplus surrogate loss and a positive class prior $\pi=0.05$, the objective is:
\begin{equation}
\mathcal{L} = \pi \widehat{R}_p^+ + \max \left( 0, \widehat{R}_u^- - \pi \widehat{R}_p^- \right)
\end{equation}
where $\widehat{R}_p^+$ and $\widehat{R}_p^-$ are empirical losses on positive data treated as positive and negative, respectively, and $\widehat{R}_u^-$ is the loss on unlabeled data treated as negative.

\section{Experiments}
\paragraph{Benchmarked Methods.}
We selected methods from three categories. First, end-to-end FIQA methods with Grad-CAM \cite{selvaraju2017grad}: MCF-Net \cite{fu2019eyeq} and FGR-Net~\cite{khalid2024fgr}. Second, explainable FIQA methods: EyeQual \cite{costa2017eyequal} and EFIQA \cite{wang2026efiqa}. Lastly, inspired by the use of Anomaly Detection (AD) techniques in EFIQA, we also trained and evaluated two AD methods: PatchCore~\cite{roth2022patchcore} and UniVAD~\cite{gu2025univad}. These models are specifically designed to identify deviations from a ``normal'' appearance without the need for explicit labels, thus offering a potential solution for label-free FIQA.

\paragraph{Training Details.}
To keep FunPiQ solely for evaluation, all the methods that do not provide pretrained weights were trained on other datasets. End-to-end methods were trained on EyeQ, as proposed in the original papers~\cite{fu2019eyeq,khalid2024fgr}, and the saliency map of the class \textit{bad} was used as quality map. For EyeQual, we reimplemented the method using a stronger backbone (ConvNeXt-tiny \cite{liu2022convnext}) and training dataset (MSHF) to ensure a fair comparison with more recent methods. Both modifications yielded significant performance improvements. AD methods were trained on a subset of the 1000 images dataset~\cite{cen20211000images} composed of pristine images. For MCF-Net and EFIQA official weights were used. To train the EFIQA-CP, we used an internal dataset consisting of 4064 CFP images from a single center, captured using a tabletop device (Topcon Maestro 2) with mixed quality.
For training, we use AdamW \cite{adamw} with a learning rate of $1\times10^{-3}$, 20 epochs, batch size of $16$, and input resolution of $1024\times1024$, with no data augmentation.

\paragraph{Evaluation Metrics.}
Given the ordinality of labels and that all models output continuous pixel-level scores, we evaluate performance using Quadratic Weighted Kappa (QWK) \cite{cohen1968qwk} and the mean S{\o}rensen--Dice coefficient (mDice) \cite{dice1945measures}. QWK measures the agreement between predicted and ground truth (GT) masks while accounting for the severity of disagreements in ordinal tasks. On the other hand, mDice, defined as the mean of the Dice scores for each class, measures the overlap between the predictions and the GT. In clinical applications, identifying bad regions is of primary interest for quality assurance. Consequently, we also evaluate performance on a binary ``Reject'' task (\textit{good}+\textit{usable} vs. \textit{bad}). We report Dice, Area Under the Receiver Operating Characteristic (AUROC) and Precision-Recall Curves (AUPRC), and sensitivity at 95\% specificity (S95), using \textit{bad} as the positive class. To ensure a fair comparison, we report the optimal results achieved by each method across a range of thresholds. Statistical significance was assessed using bootstrapping with $10,000$ iterations and a sample size of $300$ per replicate with replacement.

\section{Results}

\subsection{Quantitative Results}

\begin{table}[tb]
    \centering
    \caption{Comparison of quality assessment models in FunPiQ. Cell backgrounds indicate relative performance using min-max normalization within each dataset and metric (darker is better). Best results are in \textbf{bold}; second best, \underline{underlined}.
    }
    \label{tab:results}
    \resizebox{\textwidth}{!}{%
    \begin{tabular}{l *{6}{w{c}{1.2cm}} | *{6}{w{c}{1.2cm}}}
        \toprule
        & \multicolumn{6}{c}{\textbf{BRSET}} & \multicolumn{6}{c}{\textbf{EyeQ}} \\
        \cmidrule(lr){2-7} \cmidrule(lr){8-13}
        \textbf{Method} & \multicolumn{2}{c}{\textbf{3-Class}} & \multicolumn{4}{c}{\textbf{Reject}} & \multicolumn{2}{c}{\textbf{3-Class}} & \multicolumn{4}{c}{\textbf{Reject}} \\
        \cmidrule(lr){2-3} \cmidrule(lr){4-7} \cmidrule(lr){8-9} \cmidrule(lr){10-13}
        & QWK & mDice & Dice & AUROC & AUPRC & S95 & QWK & mDice & Dice & AUROC & AUPRC & S95 \\
        \midrule
        MCF-Net & \cellcolor{gray!22}39.54 & \cellcolor{gray!16}48.13 & \cellcolor{gray!11}28.61 & \cellcolor{gray!26}80.31 & \cellcolor{gray!13}26.49 & \cellcolor{gray!13}22.98 & 9.73 & 35.58 & \cellcolor{gray!5}62.35 & \cellcolor{gray!2}54.66 & \cellcolor{gray!4}50.57 & \cellcolor{gray!2}8.28 \\
        FGR-Net & \cellcolor{gray!4}19.16 & 39.06 & \cellcolor{gray!3}20.60 & \cellcolor{gray!4}62.19 & 12.28 & 5.48 & \cellcolor{gray!3}14.48 & \cellcolor{gray!2}37.16 & \cellcolor{gray!3}60.86 & 51.90 & 45.47 & 3.69 \\
        PatchCore & 13.95 & 38.86 & \cellcolor{gray!2}19.95 & 58.18 & \cellcolor{gray!2}15.35 & \cellcolor{gray!5}12.30 & \cellcolor{gray!8}23.40 & \cellcolor{gray!5}39.91 & 58.83 & \cellcolor{gray!10}62.92 & \cellcolor{gray!13}60.49 & \cellcolor{gray!7}15.42 \\
        UniVAD & \cellcolor{gray!3}17.71 & 39.29 & 17.17 & \cellcolor{gray!5}63.10 & \cellcolor{gray!4}16.93 & \cellcolor{gray!6}14.38 & \cellcolor{gray!1}11.81 & 36.12 & \cellcolor{gray!6}62.92 & \cellcolor{gray!4}56.58 & \cellcolor{gray!5}51.49 & \cellcolor{gray!2}7.43 \\
        EyeQual & \cellcolor{gray!32}\underline{51.32} & \cellcolor{gray!27}53.94 & \cellcolor{gray!24}40.58 & \cellcolor{gray!32}85.42 & \cellcolor{gray!22}37.33 & \cellcolor{gray!21}33.18 & \cellcolor{gray!34}64.89 & \cellcolor{gray!34}61.58 & \cellcolor{gray!28}76.39 & \cellcolor{gray!30}83.17 & \cellcolor{gray!26}76.18 & \cellcolor{gray!9}19.63 \\
        EFIQA & \cellcolor{gray!25}43.50 & \cellcolor{gray!27}\underline{54.07} & \cellcolor{gray!29}\underline{46.10} & \cellcolor{gray!32}\underline{85.45} & \cellcolor{gray!39}\underline{55.04} & \cellcolor{gray!36}\underline{52.23} & \cellcolor{gray!35}\underline{65.27} & \cellcolor{gray!34}\underline{62.09} & \cellcolor{gray!34}\underline{79.67} & \cellcolor{gray!36}\underline{89.27} & \cellcolor{gray!38}\underline{89.43} & \cellcolor{gray!35}\underline{61.70} \\
        \textit{EFIQA-CP} & \cellcolor{gray!40}\textbf{60.17} & \cellcolor{gray!40}\textbf{60.68} & \cellcolor{gray!40}\textbf{56.12} & \cellcolor{gray!40}\textbf{92.10} & \cellcolor{gray!40}\textbf{55.95} & \cellcolor{gray!40}\textbf{56.98} & \cellcolor{gray!40}\textbf{72.94} & \cellcolor{gray!40}\textbf{66.04} & \cellcolor{gray!40}\textbf{83.21} & \cellcolor{gray!40}\textbf{92.68} & \cellcolor{gray!40}\textbf{91.00} & \cellcolor{gray!40}\textbf{69.23} \\
        \midrule
        & \multicolumn{6}{c}{\textbf{mBRSET}} & \multicolumn{6}{c}{\textbf{\textit{Merged}}} \\
        \cmidrule(lr){2-7} \cmidrule(lr){8-13}
        \textbf{Method} & \multicolumn{2}{c}{\textbf{3-Class}} & \multicolumn{4}{c}{\textbf{Reject}} & \multicolumn{2}{c}{\textbf{3-Class}} & \multicolumn{4}{c}{\textbf{Reject}} \\
        \cmidrule(lr){2-3} \cmidrule(lr){4-7} \cmidrule(lr){8-9} \cmidrule(lr){10-13}
        & QWK & mDice & Dice & AUROC & AUPRC & S95 & QWK & mDice & Dice & AUROC & AUPRC & S95 \\
        \midrule
        MCF-Net & \cellcolor{gray!13}21.97 & \cellcolor{gray!8}39.40 & \cellcolor{gray!3}40.09 & \cellcolor{gray!10}62.06 & \cellcolor{gray!2}29.95 & 8.25 & \cellcolor{gray!12}31.18 & \cellcolor{gray!8}44.14 & \cellcolor{gray!7}47.05 & \cellcolor{gray!9}65.47 & \cellcolor{gray!4}35.83 & \cellcolor{gray!3}10.31 \\
        FGR-Net & \cellcolor{gray!8}13.73 & \cellcolor{gray!7}38.53 & 37.24 & \cellcolor{gray!4}55.49 & 26.80 & 8.22 & \cellcolor{gray!5}20.79 & \cellcolor{gray!4}41.01 & \cellcolor{gray!2}42.63 & \cellcolor{gray!1}58.87 & 30.48 & 6.06 \\
        PatchCore & \cellcolor{gray!14}23.22 & \cellcolor{gray!11}41.69 & \cellcolor{gray!4}40.72 & \cellcolor{gray!11}62.99 & \cellcolor{gray!8}36.69 & \cellcolor{gray!7}15.66 & \cellcolor{gray!9}26.55 & \cellcolor{gray!6}42.96 & \cellcolor{gray!4}44.64 & \cellcolor{gray!8}64.48 & \cellcolor{gray!8}40.50 & \cellcolor{gray!7}15.68 \\
        UniVAD & 0.78 & 33.90 & 37.47 & 51.73 & 27.76 & \cellcolor{gray!2}11.15 & 13.59 & 38.33 & 40.83 & 57.63 & \cellcolor{gray!2}33.42 & \cellcolor{gray!4}11.58 \\
        EyeQual & \cellcolor{gray!40}\textbf{62.43} & \cellcolor{gray!38}\underline{59.03} & \cellcolor{gray!28}57.80 & \cellcolor{gray!31}81.65 & \cellcolor{gray!18}47.92 & \cellcolor{gray!8}17.34 & \cellcolor{gray!38}\underline{66.54} & \cellcolor{gray!35}62.42 & \cellcolor{gray!30}65.92 & \cellcolor{gray!34}86.43 & \cellcolor{gray!27}64.67 & \cellcolor{gray!19}31.73 \\
        EFIQA & \cellcolor{gray!35}55.69 & \cellcolor{gray!36}57.68 & \cellcolor{gray!38}\underline{65.27} & \cellcolor{gray!38}\underline{87.96} & \cellcolor{gray!40}\textbf{73.64} & \cellcolor{gray!40}\textbf{49.24} & \cellcolor{gray!36}63.26 & \cellcolor{gray!36}\underline{62.67} & \cellcolor{gray!37}\underline{71.36} & \cellcolor{gray!37}\underline{89.45} & \cellcolor{gray!40}\textbf{80.01} & \cellcolor{gray!40}\textbf{59.01} \\
        \textit{EFIQA-CP} & \cellcolor{gray!37}\underline{59.12} & \cellcolor{gray!40}\textbf{60.06} & \cellcolor{gray!40}\textbf{66.13} & \cellcolor{gray!40}\textbf{89.32} & \cellcolor{gray!32}\underline{64.40} & \cellcolor{gray!31}\underline{40.32} & \cellcolor{gray!40}\textbf{68.41} & \cellcolor{gray!40}\textbf{65.10} & \cellcolor{gray!40}\textbf{73.45} & \cellcolor{gray!40}\textbf{91.30} & \cellcolor{gray!36}\underline{75.83} & \cellcolor{gray!37}\underline{55.61} \\
        \bottomrule
    \end{tabular}%
    }
\end{table}

The quantitative benchmark results are summarized in Table~\ref{tab:results}. Overall, EFIQA-CP achieves the best performance, followed by EFIQA and EyeQual, suggesting that explainable-by-design approaches are particularly effective for this task. EFIQA-CP ranks first on all metrics in BRSET and EyeQ. On mBRSET, it remains the top-performing method in terms of mDice and Reject Dice. On the merged dataset, compared with EFIQA, EFIQA-CP yields statistically significant improvements in QWK ($p < 0.0001$), mDice ($p < 0.01$), and Reject Dice ($p < 0.05$). Furthermore, EFIQA-CP significantly outperforms all other methods across these three metrics (all $p < 0.0001$). Interestingly, although EFIQA is the overall second-best method, EyeQual consistently surpasses EFIQA in QWK but not in mDice. EyeQual tends to confuse \textit{usable} and \textit{bad} more, with lower performance for Reject, but has better precision in the \textit{good} class, making the overall QWK better. Among the three strongest methods, their performance becomes closer in mBRSET, the most challenging dataset; notably, EFIQA attains the best AUPRC and S95. Since mBRSET is collected with mobile devices, this observation suggests that EFIQA may be more robust to domain shift.

Among the remaining methods, MCF-Net and PatchCore perform relatively well overall. MCF-Net is stronger on BRSET, whereas PatchCore performs better on EyeQ and mBRSET. The performance of general AD methods falls short of the expectations set by their success on industrial and medical benchmarks. This indicates that modern AD methods may be inherently biased toward localizing compact anomalies, struggling to capture larger quality degradations. We believe that domain-specific design is still required to fully exploit AD for FIQA.
Overall, EFIQA-CP is the most stable method across datasets and metrics.

\subsection{Qualitative Results}

\begin{figure}[h!]
    \centering
    \includegraphics[width=\linewidth]{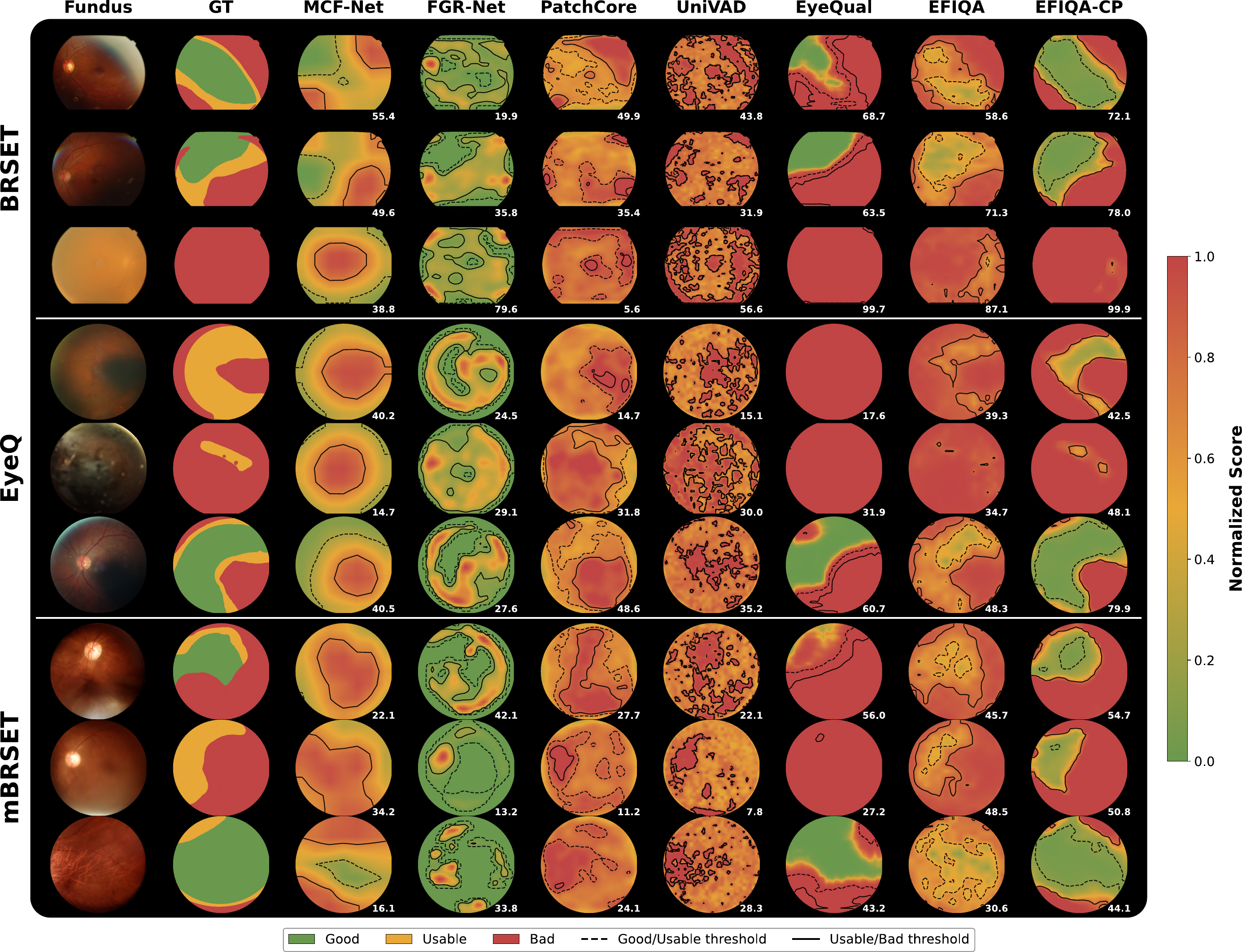}
    \caption{Qualitative comparison of quality maps from each method. Lower-right numbers report per-image mDice (\%), averaged over present ground-truth classes.}
    \label{fig:results}
\end{figure}

Qualitative results are shown in Fig.~\ref{fig:results}. We observe that EFIQA-CP produces the most precise maps, with smooth regions and a better-utilized score range. In line with the quantitative results, the three explainable-by-design methods (EFIQA-CP, EFIQA, and EyeQual) consistently yield meaningful maps across all examples. EyeQual localizes well \emph{usable} and \emph{bad} areas, but its outputs appear strongly binarized, making \emph{usable} vs.\ \emph{bad} separation less distinct. EFIQA shows better separability, while the \emph{good}--\emph{usable} transition appears less accurate and the maps are noisier than EFIQA-CP.
%
For the baselines, we observe that MCF-Net often produces plausible saliency maps but does not fully cover globally degraded images (see Fig.~\ref{fig:results}, 3\textsuperscript{rd} row). FGR-Net appears less reliable and triggers frequent false positives. PatchCore and UniVAD highlight degraded regions, but their scores remain high across patches and the maps appear noisy. Overall, explainable-by-design methods provide better quality maps, and EFIQA-CP shows a clear advantage in spatial precision and thus practical usability.

\subsection{Ablation Study}
We conducted an ablation study to assess the impact of our architectural changes and nnPU learning on final performance. All models were trained on the same in-house dataset. The model using our new architecture but trained with BCE achieved 63.36 mDice, 67.00 QWK and 90.76 AUROC, outperforming EFIQA (62.67 mDice, 63.26 QWK, 89.45 AUROC) but not the full EFIQA-CP (65.10 mDice, 68.41 QWK, 91.30 AUROC). Notably, while the BCE-trained model suffered from overfitting and performance degradation during extended training, nnPU learning ensured stable convergence.

\section{Conclusions}
In this work, we introduced FunPiQ, the first dataset with pixel-level quality annotations for evaluating FIQA models and their explanations in a task-agnostic way. In addition, we proposed EFIQA-CP, a method for precise identification of degradations that leverages a wide-yet-shallow feature adapter and nnPU learning to mitigate the inherent noise of quality pseudo-labels. Our extensive evaluation in FunPiQ revealed that explainable-by-design methods hold a distinct advantage for local quality prediction, with EFIQA-CP achieving state-of-the-art performance.
In the future, we plan to extend FunPiQ with more samples and multi-reader annotations to quantify inter-annotator agreement.
In addition, the combination of quality maps with anatomical localization to enable dynamic image-level quality criteria represents an interesting research direction.
Beyond benchmarking explainability, pixel-level maps also have direct application value: by showing \emph{where} quality is insufficient rather than returning an opaque global score, they support quality assurance for non-expert operators and can guide targeted image reacquisition.
In addition, they could provide a strong prior for CFP enhancement, by distinguishing anatomy from artifacts, helping restoration models enhance recoverable regions while avoiding hallucinations.
We believe FunPiQ benchmark will encourage research on pixel-level CFP quality detection and EBD methods, and facilitate the development of more robust and flexible FIQA methods with stronger clinical applicability.

\begin{credits}
\subsubsection{\ackname} Funded by the European Union (ERC, HealthAEye, 101171183). Views and opinions expressed are however those of the author(s) only and do not necessarily reflect those of the European Union or the European Research Council. Neither the European Union nor the granting authority can be held responsible for them. This work was also supported in part by the Christian Doppler Research Association, Austrian Federal Ministry of Economy, Energy and Tourism, and the National Foundation for Research, Technology.

\subsubsection{\discintname}
The authors have no competing interests to declare that are relevant to the content of this article.
\end{credits}


\bibliographystyle{splncs04}
\bibliography{reference}

\end{document}